# Scalable Dense Monocular Surface Reconstruction

Mohammad Dawud Ansari[1]
md_dawud.ansari@dfki.de

Vladislav Golyanik[1,2]
vladislav.golyanik@dfki.de

Didier Stricker[1,2]
didier.stricker@dfki.de

[1] Department of Augmented Vision, DFKI   [2] University of Kaiserslautern

## Abstract

*This paper reports on a novel template-free monocular non-rigid surface reconstruction approach. Existing techniques using motion and deformation cues rely on multiple prior assumptions, are often computationally expensive and do not perform equally well across the variety of data sets. In contrast, the proposed Scalable Monocular Surface Reconstruction (SMSR) combines strengths of several algorithms, i.e., it is scalable with the number of points, can handle sparse and dense settings as well as different types of motions and deformations. We estimate camera pose by singular value thresholding and proximal gradient. Our formulation adopts alternating direction method of multipliers which converges in linear time for large point track matrices. In the proposed SMSR, trajectory space constraints are integrated by smoothing of the measurement matrix. In the extensive experiments, SMSR is demonstrated to consistently achieve state-of-the-art accuracy on a wide variety of data sets.*

## 1. Introduction

The class of methods aiming at 3D surface reconstruction by factorising given 2D point tracks across monocular views is known as non-rigid structure from motion (NRSfM). Monocular surface recovery has been tackled for perspective and orthographic cases. We assume an uncalibrated set of views and adopt orthographic camera model in this paper. Unlike its rigid counterpart [32], NRSfM is a highly ill-posed problem with several inherent ambiguities such as basis, global rotation and reflection ambiguities. Lots of research has been done in this field so far, and it remains very active [35, 33, 3, 28, 19, 12, 23].

One difficulty in NRSfM arises due to the basis ambiguity [35]. Most of the current approaches use additional constraints to handle this problem. Many methods assume the motion of the camera to be small and smooth [33, 3, 28, 19, 12] which limits their applicability in practice. Another assumption found in the literature is smoothness prior on point trajectories [3] which, however, often results in oversmoothing of sudden structure changes. Besides, only several methods can handle dense sequences so far [29, 12, 15, 1, 16]. and only a few data sets for quantitative evaluation of NRSfM methods are currently available.

With these limitations in mind, our motivation is to find a scalable NRSfM method, *i.e.*, a method supporting point track matrices of different sizes, to make NRSfM robust to self-occlusions and easy to implement. Among existing approaches, it is hard to find one which fulfils all the properties stated above. Accordingly, our *contributions* in this paper can be summarised as follows:

- We design an NRSfM approach which is easy to implement, computationally efficient, can potentially run interactively and can be used for a variety of realistic data sets. We call the new method Scalable Monocular Surface Reconstruction, or SMSR (Sec. 3). In our method, optimal basis selection is performed automatically which reduces the overall number of unknowns. Moreover, a non-convex formulation for camera pose estimation is solved iteratively using singular value thresholding (SVT) and proximal gradient. Surface estimation is formulated as an optimisation problem where guaranteed convergence to global minima is achieved using alternating direction method of multipliers (ADMM) [25, 7].

- A novel pre-processing step for input matrix is proposed, *i.e.*, the measurement matrix is reinitialised with smooth trajectory constraints. This provides consistently accurate shape recovery for a wide variety of data sets, both sparse and dense (Sec. 3.2).

- Due to lack of dense data sets with ground truth for quantitative evaluation, we propose a two-step scheme. First, 3D shapes are estimated using an accurate and computationally expensive Variational Approach (VA) [12] followed by applying a rotation pattern in the second step (Sec. 4.1). Eventually, we found realistic data sets which allow to demonstrate strengths of the method compared to the state-of-the-art.

- We perform an extensive evaluation of SMSR on a variety of sparse and dense data sets, both synthetic and real ones (Sec. 4). Across various evaluation scenarios, SMSR achieves *consistent state-of-the-art level accuracy* compared to the competing approaches.

## 2. Related Work

The first NRSfM approach based on the Low-Rank Shape Model (LRSM) was proposed by Bregler *et al*. [8] as an extension of the Tomasi-Kanade factorisation based rigid structure from motion [32]. In LRSM, every non-rigid shape is represented by a linear combination of unknown basis shapes. The basis ambiguity has plagued early NRSfM approaches. Despite the basis and coefficients are ambiguous, it was proven that the shapes can be, nevertheless, reconstructed uniquely [2]. Dai *et al*. [10] proposed a method with a minimal number of prior assumptions but the performance of their method considerably varies with the data set. Methods adopting trajectory space model (TSM) allowed to increase stability and accuracy of monocular surface recovery due to a trajectory basis defined a priori (*e.g.*, discrete cosine transform or DCT basis) [4, 19, 21]. A predefined basis is, at the same time a disadvantage, and the performance may likewise vary considerably depending on data. The first approach that used constraints in both metric and trajectory space was [19]. Russell *et al*. [29] demonstrated dense monocular surface recovery for the first time. The authors showed how existing piecewise NRSfM techniques could be improved for scalability with the number of points. Several methods allow determining an optimal basis automatically [12, 24, 15]. Garg *et al*. [12] proposed VA with a total variation (TV) spatial regulariser. VA achieves high accuracy for multiple data sets but strongly depends on initialisation and requires an additional connectivity data structure thereby increasing the number of inputs.

A class of methods aiming at the reconstruction of scenes with large deformations adopts the idea of piecewise processing. One of the recent methods by Lee *et al*. [23] achieves state-of-the-art accuracy on several sparse data sets. However, it scales poorly with the number of points and involves several computationally expensive steps. In [10], non-convex energy minimisation is performed directly by relaxing the matrix rank constraint to a nuclear norm using semidefinite programming solver.

Our method is an algorithmic mixture of several concepts encountered in several approaches. We believe that dense NRSfM methods should be well scalable with the number of points, *i.e.*, perform equally well and stable on a broad range of data sets, with small and large deformations. Thus, we adopt constraints in the metric and trajectory spaces and our formulation is most closely related to [23]. Compared to [19], we apply trajectory space constraints for smoothing of the measurement matrix and not throughout the whole optimisation. As a distinctive characteristic, we use proximal gradient method for optimisation [5]. To find the final shapes, we coarsely follow Block Matrix Method (BMM) [10]. Instead of using a fixed point continuation algorithm to relax rank minimisation to a nuclear norm, we use ADMM, similar to the patch reconstruction step of Lee *et al*. [23]. In contrast to the latter method, our approach is not piecewise. One major advantage of using ADMM is a guaranteed global minimum for a convex optimisation problem. In our experiments, this property has been proven to be effective for the global dense setting.

## 3. Proposed Approach

SMSR consists of two main steps. First, the camera motion is estimated followed by 2D shape alignment. Next, a time-varying 3D surface is recovered. In the following, we describe the proposed algorithmic combination.

### 3.1. Problem Formulation

We formulate NRSfM as a factorisation approach assuming every non-rigid shape follows LRSM. The measurement matrix $\mathbf{W} \in \mathbb{R}^{2T \times N}$ combining coordinates of $N$ tracked points thoughout $T$ frames is decomposed as a product of two matrices:

$$\underbrace{\begin{bmatrix} \mathbf{W}_1 \\ \mathbf{W}_2 \\ \vdots \\ \mathbf{W}_T \end{bmatrix}}_{\mathbf{W}} = \underbrace{\begin{bmatrix} \mathbf{R}_1 & & & \\ & \mathbf{R}_2 & & \\ & & \ddots & \\ & & & \mathbf{R}_T \end{bmatrix}}_{\mathbf{R}} \underbrace{\begin{bmatrix} \mathbf{S}_1 \\ \mathbf{S}_2 \\ \vdots \\ \mathbf{S}_T \end{bmatrix}}_{\mathbf{S}}, \quad (1)$$

where $\mathbf{R}$ is a quasi-block-diagonal matrix of camera poses, $\mathbf{S}$ is the block matrix with non-rigid shapes, $\mathbf{R}_i \in \mathbb{R}^{2 \times 3}$ and $\mathbf{S}_i \in \mathbb{R}^{3 \times N}$ are camera pose and shape for $i^{th}$ frame respectively. Every frame in $\mathbf{W}$ is registered to the centroid of the observed structure. Alternatively, the measurement matrix can be written as

$$\mathbf{W} = \mathbf{R} \underbrace{(\mathbf{C} \otimes \mathbf{I}_3) \mathbf{B}}_{\mathbf{S}} = \mathbf{MB}, \quad (2)$$

where $\otimes$ denotes Kronecker product, $\mathbf{I}_3$ is a $3 \times 3$ identity matrix, $\mathbf{B} \in \mathbb{R}^{3K \times N}$ denotes set of basis shapes of cardinality $K$ and $\mathbf{C} \in \mathbb{R}^{T \times K}$ is the corresponding *coefficient matrix*; $\mathbf{M} \in \mathbb{R}^{2T \times 3K}$ is a combined block diagonal shape-coefficient matrix (an element of the motion manifold). We utilise singular value decomposition to find a $3K$-rank approximation of $\mathbf{W}$ as $\mathbf{W} \cong \mathbf{M}'\mathbf{B}'$. An "implicit" non-unique solution can be obtained up to an invertible matrix $\mathbf{Q} \in \mathbb{R}^{3K \times 3K}$ (corrective transformation):

$$\mathbf{M}'\mathbf{Q} \cong \mathbf{M}, \quad \mathbf{Q}^{-1}\mathbf{B}' \cong \mathbf{B}. \quad (3)$$

It can be noticed that $\mathbf{M}'\mathbf{Q}$ is a scaled orthogonal matrix. Let $\mathbf{M}'_{2i-1:2i} \in \mathbb{R}^{2 \times 3K}$ denote the $i^{th} \in \{1, \ldots, T\}$ pair of rows of $\mathbf{M}$ and $\mathbf{Q}_k \in \mathbb{R}^{3K \times 3}$ denote the $k^{th} \in \{1, \ldots, K\}$ column triplet of $\mathbf{Q}$. Then, for every $\mathbf{M}'_{2i-1:2i}$ submatrix, the following product holds:

$$\mathbf{M}'_{2i-1:2i}\mathbf{Q}_k = c_{ik}\mathbf{R}_i, \quad (4)$$

where $c_{ik}$ is the element in $i^{th}$ row and $k^{th}$ column of $\mathbf{C}$. Let $\mathbf{F} = \mathbf{Q}\mathbf{Q}^\mathsf{T}$ be a positive semi-definite matrix. Using orthogonality constraint, two systems of linear equations can be obtained for each $i$:

$$\begin{cases} \mathbf{M}'_{2i-1}\mathbf{F}_k\mathbf{M}'^T_{2i-1} = \mathbf{M}'_{2i}\mathbf{F}_k\mathbf{M}'^T_{2i} = c_{ik}^2\mathbf{I}, \\ \mathbf{M}'_{2i-1}\mathbf{F}_k\mathbf{M}'^T_{2i} = 0. \end{cases} \quad (5)$$

The system of equations (5) can be rewritten as

$$\underbrace{\begin{bmatrix} \mathbf{M}'_{2i-1} \otimes \mathbf{M}'^T_{2i-1} - \mathbf{M}'_{2i} \otimes \mathbf{M}'^T_{2i} \\ \mathbf{M}'_{2i-1} \otimes \mathbf{M}'^T_{2i} \end{bmatrix}}_{\mathbf{A}_i} \operatorname{vec}(\mathbf{F}_k) = 0, \quad (6)$$

where $\operatorname{vec}(\cdot)$ is vectorisation operator defined as $\operatorname{vec}(\mathbb{R}^{m \times n}) \Rightarrow \mathbb{R}^{mn \times 1}$. Here, we use the property:

$$\operatorname{vec}(\mathbf{\Lambda}\mathbf{\Psi}\mathbf{\Upsilon}^T) = (\mathbf{\Upsilon} \otimes \mathbf{\Lambda})\operatorname{vec}(\mathbf{\Psi}), \quad (7)$$

which holds for real matrices $\mathbf{\Lambda}$, $\mathbf{\Upsilon}$ and $\mathbf{\Psi}$. We denote the matrix on the left side of Eq. (6) with $\mathbf{A}_i$. Assembling equations for all $\mathbf{A}_i$ by stacking leads to a single equation

$$\mathbf{A}\operatorname{vec}(\mathbf{F}_k) = 0, \quad (8)$$

where $\mathbf{A} = [\mathbf{A}_1^T, \mathbf{A}_2^T, \ldots, \mathbf{A}_T^T]$. The optimisation problem in Eq. (8) — finding an optimal $\mathbf{F}_k$ — can be solved with linear least-squares by minimising

$$\|\mathbf{A}\operatorname{vec}(\mathbf{F}_k)\|^2 \quad (9)$$

[27]. The latter problem is non-convex due to the rank-3 constraint on $\mathbf{F}_k$, and can be efficiently solved using proximal gradient method [5] in an iterative manner. In each iteration, we solve the following subproblem:

$$\min_{\mathbf{F}_k} \|\mathbf{F}_k - \mathbf{F}_k^*\|^2, \text{ s.t } \operatorname{rank}(\mathbf{F}_k) = 3, \quad (10)$$

$$\operatorname{vec}(\mathbf{F}_k^*) = \operatorname{vec}(\mathbf{F}_k^0) - \frac{1}{L_\mathbf{A}}\mathbf{A}^\mathsf{T}\mathbf{A}\operatorname{vec}(\mathbf{F}_k^0). \quad (11)$$

In Eq. (10), $\mathbf{F}_k^0$ is an initial estimate of $\mathbf{F}_k$ and $L_\mathbf{A}$ denotes Lipschitz constant, *i.e.*, the largest eigenvalue of $\mathbf{A}^\mathsf{T}\mathbf{A}$. Once an optimal $\mathbf{F}$ is found, the corrective transformation $\mathbf{Q}$ is recovered by Cholesky decomposition. Further, $\mathbf{R}$ is computed using Eq. (4) and (5) with a known $\mathbf{Q}$.

### 3.2. Smooth Shape Trajectory

For an enhanced accuracy, we add smoothness constraint on point trajectories. It enforces the non-rigid shapes to change gradually over time, *i.e.*, a trajectory represented by a $K$-dimensional $t^{th} \in \{1,..,T\}$ vector $c_t|_{1:K} = c(t)$ must lie in a linear subspace which constraints a point $c_{t,k}$ to vary smoothly over time. Thus, we represent $\mathbf{C}$ by $K$ compact cosine series:

$$\mathbf{C} = \Omega_d [x_1, \quad \ldots \quad , x_K] = \Omega_d \mathbf{X} \quad \text{with} \quad x_k \in \mathbb{R}^d, \quad (12)$$

where $k \in \{1,..,K\}$ and $d$ are numbers of low frequency DCT coefficients in the shape trajectory $\mathbf{X} \in \mathbb{R}^{d \times K}$ which represents $\mathbf{C}$ compactly over a truncated DCT domain $\Omega_d \in \mathbb{R}^{T \times d}$. The advantage of the constraints in the trajectory space is a known DCT basis which reduces considerably the number of unknowns. This leads to a faster overall convergence. Using the representation in trajectory space, $\mathbf{M}$ can be written as

$$\mathbf{M} = \mathbf{R}(\mathbf{C} \otimes \mathbf{I}_3) = \mathbf{R}(\Omega_d \mathbf{X} \otimes \mathbf{I}_3). \quad (13)$$

Using the pre-computed camera motion $\mathbf{R}$ (see **Sec.** 3.1), $\mathbf{M}$ is treated as a function of $\mathbf{X}$ only. Further, $\mathbf{B}$ can be computed as

$$\mathbf{B} = \mathbf{M}^\dagger \mathbf{W}, \quad (14)$$

whereby $\dagger$ denotes Moore-Penrose pseudo-inverse operator [14]. The problem of estimating DCT shape trajectory $\mathbf{X}$ is formulated as a problem of minimising the squared reprojection error in the Frobenius norm:

$$\min_{\mathbf{M}} \|\mathbf{W} - \mathbf{W}^*\|_\mathcal{F}^2, \quad \text{and} \quad \mathbf{W}^* = \mathbf{MB} = \mathbf{MM}^\dagger\mathbf{W}. \quad (15)$$

$\mathbf{X}$ is initialised to $\mathbf{X}_0 = [\mathbf{I}_K \ 0]$ which is $K \times K$ identity matrix with padding of additional zeros. The higher frequency DCT coefficients can be then estimated by iterative Gauss-Newton minimisation of Eq. (15). Shape basis matrix $\mathbf{B}$ is not directly used in the next part of the pipeline. Instead, only a 2D projection obtained from the newly computed matrix $\mathbf{S} = (\Omega_d \mathbf{X} \otimes \mathbf{I}_3)\mathbf{B}$ is used as the new measurement matrix $\mathbf{W}$. The described pre-processing step reinitialises the input matrix after imposing the smooth trajectory constraint on the recovered shapes.

### 3.3. Non-rigid Shape Recovery

We use the new $\mathbf{W}$ to recover the final shapes and follow the formulation proposed in [10]. A rearranged shape matrix

$$\mathbf{S}^\# = \begin{bmatrix} X_{11}\ldots X_{1N} & Y_{11}\ldots Y_{1N} & Z_{11}\ldots Z_{1N} \\ \vdots & \vdots & \vdots & \vdots & \vdots & \vdots \\ X_{T1}\ldots X_{TN} & Y_{T1}\ldots Y_{TN} & Z_{T1}\ldots Z_{TN} \end{bmatrix} \quad (16)$$

with an additional constraint $\operatorname{rank}(\mathbf{S}^\#) < K$ compactly represents the optimal non-rigid structure:

$$\min_{\mathbf{S}} \|\mathbf{S}^\#\|_*, \quad \text{s.t.} \quad \mathbf{W} = \mathbf{RS}, \quad (17)$$

where $||.||_*$ denotes the nuclear norm. We assume the mean 3D component is dominant in $\mathbf{S}^\#$ and can be removed in temporal dimension. Eq. (17) can be modified as

$$\min_{\mathbf{S}} ||\mathbf{S}^\# \mathbf{P}||_*, \quad \text{s.t.} \quad \mathbf{W} = \mathbf{R}\mathbf{S}, \quad (18)$$

where $\mathbf{P} = (\mathbf{I} - \frac{1}{T}\mathbf{1}\mathbf{1}^T)$ is the orthogonal projection and $\mathbf{1}$ is a vector of ones being its own null space. Eq. (18) is an optimisation problem over a convex function. We use for optimisation ADMM which can be considered as the inner-loop version of augmented Lagrangian method [25]:

$$\begin{aligned} L(\mathbf{S}, \mathbf{Y}, \mu) &= \\ &= ||\mathbf{S}^\# \mathbf{P}||_* + \mathbf{Y}^T(\mathbf{W} - \mathbf{R}\mathbf{S}) + \frac{\mu}{2}||\mathbf{W} - \mathbf{R}\mathbf{S}||_\mathcal{F}^2, \end{aligned} \quad (19)$$

where $\mathbf{Y}$ denotes the vector of Lagrange multipliers and $\mu > 0$ is the positive step size. Eq. (18) is solved by the optimisation

$$\max_{\mathbf{Y}} \min_{\mathbf{S}} L(\mathbf{S}, \mathbf{Y}, \mu). \quad (20)$$

$\mathbf{S}$ is initialised as a planar structure and later updated iteratively by SVT [25] using coordinate descent. Lagrange multipliers are updated using sub-gradients:

$$\mathbf{S}^{t+1} = \arg\min_{\mathbf{S}} L(\mathbf{S}, \mathbf{Y}^t, \mu^t), \quad (21)$$

$$\mathbf{Y}^{t+1} = \mathbf{Y}^t + \mu^t(\mathbf{W} - \mathbf{R}\mathbf{S}^{t+1}), \quad (22)$$

$$\mu^{t+1} = \rho\, \mu^t, \quad (23)$$

where $\mu$ grows geometrically in every iteration $t$ with the growth rate $\rho$. One advantage of ADMM is a guaranteed global minimum for convex problems and we use the above modified formulation to estimate the final shapes.

## 4. Experimental Results

We perform experiments on both synthetic as well as real data and compare the proposed approach with several existing approaches, *i.e.*, Metric Projection (MP) [28] (for the dense data sets we also use accelerated MP [18]), Point Trajectory Approach (PTA) [3], smooth time trajectories approach by Gotardo *et al.* (CSF1) [19], complementary rank-3 spaces approach CSF2 [21], Dai *et al.* (BMM) [10], VA [12] and Lee *et al.* [23]. We run experiments on a system with 32 GB RAM and Intel Core-i5200U processor. For SMSR implementation we used *Matlab* programming tool [26]. For quantitative performance evaluation of the methods, we compute normalised mean 3D error $e_{3D}$ given by

$$e_{3D} = \frac{1}{\sigma T N} \sum_{t=1}^{T} \sum_{j=1}^{n} e_{tj}, \; \sigma = \frac{1}{3T} \sum_{t=1}^{T} (\sigma_{tx} + \sigma_{ty} + \sigma_{tz}), \quad (24)$$

where $\sigma$ is the normalised variance and $e_{tj}$ is the reconstruction error (*i.e.*, Euclidean distance for the $j^{th}$ 3D point of $t^{th}$ frame) between ground truth $\mathbf{G}$ and recovered shape $\mathbf{X}$ given by $e_{tj} = ||\mathbf{G}_j^t - \mathbf{X}_j^t||_\mathcal{F}^2$, where $||.||_\mathcal{F}^2$ is the squared Frobenius norm. Before computing the error metrics, the Procrustes analysis is performed to align recovered shapes with the ground truth [22].

### 4.1. Dense Data Sets with Ground Truth

To generate more data sets for quantitative evaluation, we propose to employ the available dynamic 3D reconstructions. For a given input, we firstly reconstruct non-rigid shapes with VA [12]. Next, the 3D shapes are transformed to the fronto-parallel position relative to the image plane and projected onto it with a virtual orthographic camera. Thus, ground truth point correspondences are obtained. The 3D shapes and point correspondences can now be utilised as a data set for NRSfM evaluation with a known ground truth. We obtained three data sets in this way — *face* [12], *heart* [30] and *barn owl* [11]. While generating 2D measurement matrices, for all sequences we apply the recovered camera poses to the dynamic shapes.

Additionally, we use the actor data set of Beeler *et al.* with given dynamic 3D shapes [6]. The sequence was acquired using motion capture techniques and interpolation of key frame shapes; it contains $\approx 1.1 \times 10^6$ points and over 300 frames with subtle details of an actor's face with different facial expressions and emotions. We use every odd frame from the first 101 frames in the experiment and create two modified data sets, namely *Actor1* — with the introduced left-right head movement — and *Actor2* — with the introduced left-right up-down head movement. Similar to the *face*, *heart* and *barn owl* sequences, we project the 3D shapes by an orthographic camera and obtain 2D correspondences. In the similar manner — using the new sequences *Actor1* and *Actor2* — we created modified data sets *Actor1 Sparse* and *Actor2 Sparse*, as some implementations were only able to execute with less points ($\approx 3.8 \times 10^4$).

### 4.2. Evaluation on Sparse Data Sets

We evaluate SMSR on challenging synthetic data sets *Drink, Pickup, Stretch, Yoga* with added per frame angular displacement of 5 degrees [4]. They contain 41 tracked points and up to 1102 frames. Further, SMSR is evaluated on real sparse benchmark *Dinosaur* and *Dance* sequences introduced in [4].

In Table 1, we show that the performance of the proposed method attains very close to the respective best performing method for nearly all cases (for *Drink* and *Yoga* sequences, SMSR matches the best $e_{3D}$ up to the second decimal digit). Moreover, CSF1 achieves $e_{3D}$ of 0.023 and 0.602 for *Drink* sequence and *Dinosaur* sequence and our method achieves 0.028 and 0.467 for the respective sequences. Hence, SMSR shows robust behaviour and steady performance on all sparse benchmark data sets.

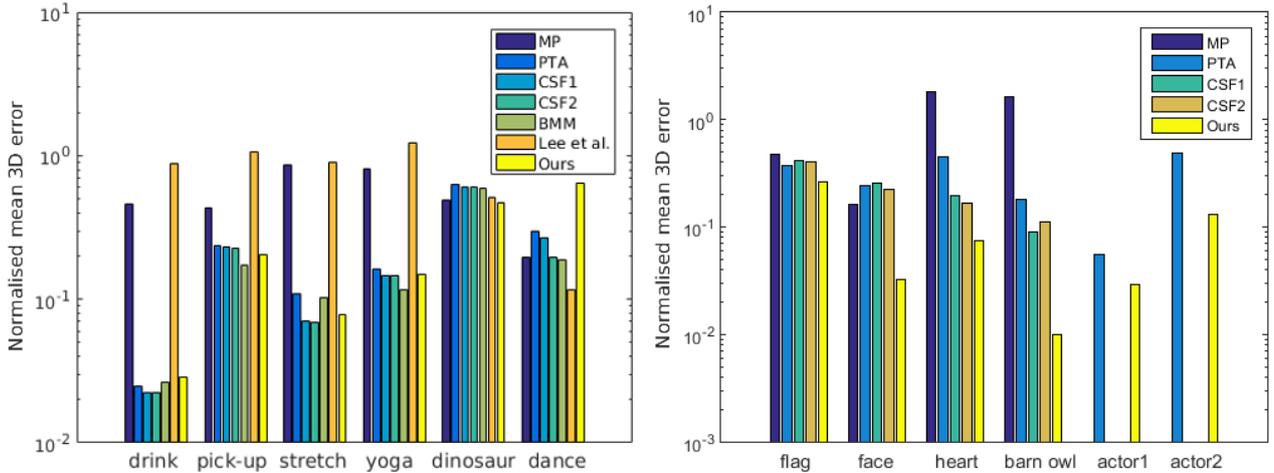

**Figure 1:** Comparison of the log scale normalised mean 3D error of SMSR (proposed) and several other methods over sparse and dense benchmark data sets. Missing bars in the figure for some approaches corresponds to N.A. in Table 2.

| Method | *Drink*(K) | *Pick-up*(K) | *Stretch*(K) | *Yoga*(K) | *Dinosaur*(K) | *Dance*(K) |
|---|---|---|---|---|---|---|
| MP [28] | 0.4604 | 0.4332 | 0.8549 | 0.8039 | **0.4894** | 0.1949 |
| PTA [3] | 0.0250(13) | 0.2369(12) | 0.1088(12) | 0.1625(11) | 0.6265(2) | 0.2958(5) |
| CSF1 [19] | **0.0223**(6) | 0.2301(6) | **0.0710**(8) | **0.1467**(7) | 0.6022(2) | 0.2694(2) |
| CSF2 [21] | **0.0223**(6) | 0.2277(3) | **0.0684**(8) | **0.1465**(7) | 0.6003(2) | 0.1957(7) |
| BMM [10] | 0.0266(12) | **0.1731**(12) | 0.1034(11) | **0.1150**(10) | 0.5874 | 0.1864(10) |
| Lee *et al.* [23] | 0.8754 | 1.0689 | 0.9005 | 1.2276 | 0.5079 | **0.1156** |
| SMSR (*ours*) | **0.0287** | **0.2020** | **0.0783** | **0.1493** | **0.4678** | 0.6407 |

**Table 1:** Normalised mean 3D error $e_{3D}$ for benchmark data sets [33, 3, 20]. $K$ denotes number of bases which has led to the best result.

### 4.3. Evaluation on Dense Data Sets

We use in the experiments the challenging dense synthetic mocap *flag* data set with ground truth surfaces [34, 13] with $\approx 10^4$ points and 60 frames. To the best of our knowledge, no real dense data set with the ground truth is available for quantitative evaluation of NRSfM yet. Aiming to fill this gap, we create further synthetic dense data sets with ground truth using publicly available sequences (see Sec. 4.1), *i.e.*, the *face* sequence with $\approx 28 \times 10^3$ points and 120 frames [12] (it contains sudden emotion changes between frames) and the *heart* bypass surgery sequence [30] with $\approx 68 \times 10^3$ points and 30 frames. We used multi-frame optical flow [31] to compute dense correspondences. To demonstrate the scalable nature of SMSR, we have tested it on a dense real *barn owl* video acquired outdoors [11] with a higher number of points and frames ($\approx 2 \times 10^5$ points, 202 frames). Additionally, we use four synthetic face sequences introduced in [12] denoted by *Face Seq1 – Face Seq4*. Each sequence contains 28887 points per frame. Note that the initial shape alignment by imposing smooth trajectories constraints (described in Sec. 3.2) takes longer time to converge where the total number of point tracks is high.

As a result, for these sequences, the initial shape alignment is disabled. Table 2 shows performance for mocap synthetic flag sequence [34] and real sequences [6, 12, 11]. The proposed method achieves the best performance and outperforms other approaches by a considerable margin. From Table 2 we can infer that other approaches exhibit high variation in performance across tested data sets and scale poorly for different deformations, number of points and frames. In contrast, SMSR performs robustly and is well scalable for sparse and dense data and different types of deformations.

In Table 3, we provide a quantitative comparison for several NRSfM methods on the newly created data sets *Actor1 Sparse*, *Actor2 Sparse* and synthetic face sequences *Face Seq1 – Face Seq4*. The $e_{3D}$ obtained by our approach remains lower than 0.04 for the modified actor data sets evidencing the consistent performance of our method. The qualitative performance evaluation for *Actor1 Sparse* sequence is shown in Fig. 2. SMSR accurately recovers actor's face with distinct fine details while most of the other approaches fail. According to Table 3, SMSR achieves second best accuracy with $e_{3D}$ being less than 0.21 while VA achieves best results for all synthetic face sequences, it is found that these sequences contain overexaggerated facial deformations (*cf.* results of VA on sequences *Actor1 Sparse* and *Actor2 Sparse* with smaller and realistic deformations). We visualise 3D motion fields on the surface following the methodology proposed in [17]. Accordingly, in Fig. 3 we show two frames per data set in the top row and the colour-coded magnitudes of the recovered 3D motion fields overlayed with the recovered shapes in the bottom row. Finally, Fig. 4 shows reconstructions obtained using various approaches — the third and fourth rows show that the SMSR reconstructions of *Face Seq1 – Face Seq4* sequences remain very close to those of VA and ground truth.

| Method | *flag*(K) | *face*(K) | *heart*(K) | *barn owl*(K) | *Actor1*(K) | *Actor2*(K) |
|---|---|---|---|---|---|---|
| MP [28] | 0.4756(8) | 0.1604(2) | 1.8231(4) | 1.6277(4) | N.A. | N.A. |
| PTA [3] | 0.3755(2) | 0.2410(3) | 0.4509(2) | 0.1805(2) | 0.0559(2) | 0.4941(3) |
| CSF1 [19] | 0.4202(4) | 0.2581(2) | 0.1941(2) | 0.0896(2) | N.A. | N.A. |
| CSF2 [21] | 0.4021(4) | 0.2226(2) | 0.1667(2) | 0.1101(2) | N.A. | N.A. |
| SMSR (*ours*) | **0.2631** | **0.0321** | **0.075** | **0.0099** | **0.0287** | **0.1318** |

**Table 2:** Normalised mean 3D error $e_{3D}$ for benchmark data sets [34, 6, 12, 11]. $K$ denotes number of bases which has led to the best result. N.A. is used where evaluation is not possible because of very high memory requirement or infeasible runtime. Likewise, BMM [10] and Lee *et al.* [23] could not be evaluated. The newly generated ground truth (see Sec. 4.1) does not contain connectivity data structure, thus running VA was not possible on these data sets.

| Method | MP [28] | PTA [3] | CSF1 [19] | CSF2 [21] | VA [12] | SMSR (*ours*) |
|---|---|---|---|---|---|---|
| Actor1 Sparse(K) | 0.5226(3) | **0.0418**(2) | 0.3711(2) | 0.3708(2) | N.A. | **0.0352** |
| Actor2 Sparse(K | 0.2737(2) | **0.0532**(2) | 0.2275(3) | 0.2279(3) | N.A. | **0.0334** |
| Face Seq1(K) | 0.7251(2) | 0.3933(2) | 0.5325(3) | 0.4677(3) | **0.1058** | 0.1893 |
| Face Seq2(K) | 0.6633(2) | **0.1871**(2) | 0.9266(3) | 0.7909(3) | **0.1014** | 0.2133 |
| Face Seq3(K) | 0.5676(4) | 0.1706(4) | 0.5274(3) | 0.5474(3) | **0.0811** | 0.1345 |
| Face Seq4(K) | 0.5038(4) | 0.2216(4) | 0.5392(4) | 0.5292(3) | **0.0806** | **0.0984** |

**Table 3:** Normalised mean 3D error $e_{3D}$ for the modified benchmark data set [6, 12]. $K$ denotes number of bases which has led to the best result. BMM [10] and Lee *et al.* [23] could not be evaluated for *Actor1 Sparse* and *Actor2 Sparse* sequences because of infeasible memory requirement and very high runtime. The results of VA are included as presented in the publication [12]. N.A. is used for the newly generated data sets (see Sec. 4.1) which does not contain connectivity data structure, thus running VA was not possible on these data sets.

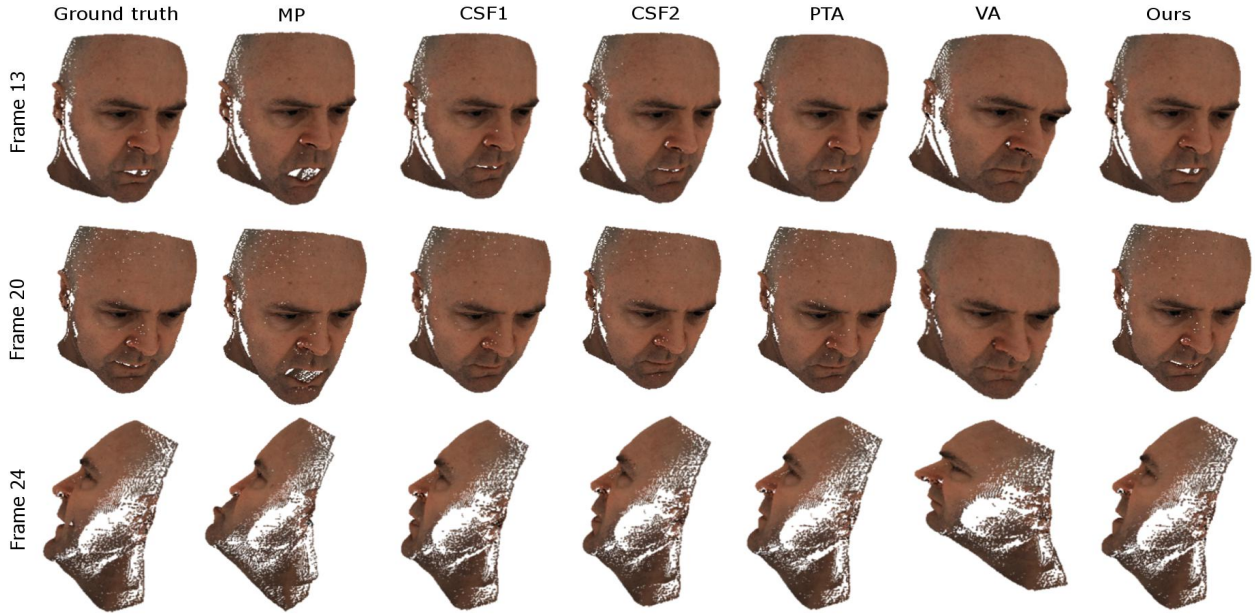

**Figure 2:** Qualitative comparison of MP [28], CSF1 [19], CSF2 [21], PTA [3], VA [12] and our SMSF approach for the *Actor1 Sparse* sequence.

SMSR requires only a few user-specified settings — the step size $\mu$ and the growth rate $\rho$. In all experiments, we set

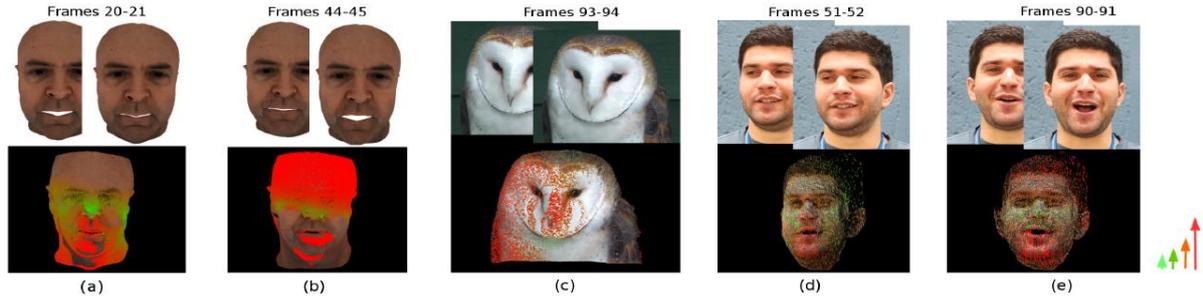

**Figure 3:** Visualisation of the 3D motion fields in the sense of the NRSfM-Flow framework [17] from the 3D shapes recovered by our SMSF method, for the *Actor* [6] (a, b), *barn owl* [11] (c) and *face* [12] (d,e) dense data sets. The scheme on the right shows relative vector lengths and corresponding colours.

$\mu = 1$ and $\rho = 1.02$. For all sparse benchmark data sets, SMSR takes less than 5 *sec* for the reconstructions, and between 40 and 300 *sec* for the dense data sets (*flag, heart, face*). Other sequences, *i.e.*, *barn owl* and *Actor* sequences can be reconstructed offline in feasible time ($\leq 2500\,sec$) because of the high number of points. Note that our current implementation is not optimised. With hardware acceleration SMSR can be suitable for interactive applications.

### 4.4. NRSfM Challenge 2017

We tested SMSR on the NRSfM Challenge 2017 benchmark[1] [9]. This benchmark represents a set of five measurement matrices corresponding to five different monocular scene observations with small frame-to-frame changes — *Articulated Joints*, *Balloon Deflation*, *Paper Bending*, *Rubber Stretching* and *Paper Tearing* — and reports average root mean square (RMS) error[2] in $mm$ for seventeen NRSfM methods with publicly available source code and methods submitted for evaluation. Measurement matrices for both perspective and orthographic cameras are provided. We reconstructed all sequences with the proposed SMSR and submitted to the portal (note that the ground truth 3D shapes are not publicly available). We use default parameters across all test sequences. In the mode without self-occlusions, SMSR achieves the average RMS error of $41.84\,mm$ (ranging from $31.71\,mm$ for *Balloon Deflation* to $58.12\,mm$ for *Articulated Joints*) with the standard deviation of $9.2\,mm$ — an intermediate overall accuracy. The low value of the standard deviation accentuates the scalable nature of our approach. SMRS outperforms Lee *et al.* [23] and Dai *et al.* [10], though [28, 21, 19, 4] achieve lower errors on the new benchmark.

---

[1] see http://nrsfm2017.compute.dtu.dk/benchmark for more details

[2] defined as $\frac{1}{T}\sum_{t=1}^{T}\frac{\|\mathbf{X}^t - \mathbf{G}^t\|_{\mathcal{F}}}{\mathbf{G}^t}$, with reconstructions $\mathbf{X}^t$ and ground truth shapes $\mathbf{G}^t$

### 5. Conclusion

This paper presents a new method for scalable NRSfM which involves two main steps — camera motion estimation and 3D surface update. The primary advantage of the proposed SMSR approach is superior scalability and consistent performance across different data sets which are two core synergic properties which distinguish our SMSR method from the competing methods. SMSR imposes constraints in metric and trajectory spaces and involves various widely-used high-level operations in different steps including proximal gradient method, iterative Gauss-Newton optimisation, singular value thresholding, coordinate descent, Cholesky decomposition as well as multiple matrix-vector and matrix multiplications. We often observed a quick convergence of our method. The reconstruction error obtained in the extensive experiments by the proposed approach is low and consistent over a large variety of publicly available real and synthetic data sets including sparse and dense ones. To facilitate and enrich quantitative evaluation, we evaluated performance of several approaches on ground truth tracks obtained as projections of trail VA reconstructions. In the majority of the cases, SMSR achieves the lowest normalised mean 3D error among all tested methods or reaches close to the respective best performing method. One reason for this gain is the new pre-processing step, *i.e.*, smoothing of the shape trajectory. Another possible reason is the convergence property of the ADMM minimisation.

A current limitation of the proposed technique lies in handling large deformations such as those occurring during tissue bending and twisting. Extending SMSR for the handling of large deformations is an interesting direction for the future work. Another possible direction could be adopting the method for online applications and embedded devices.

### Acknowledgements

This research was supported by the project DYNAMICS (01IW15003) of the German Federal Ministry of Education and Research (BMBF). We thank Gerd Reis for comments.

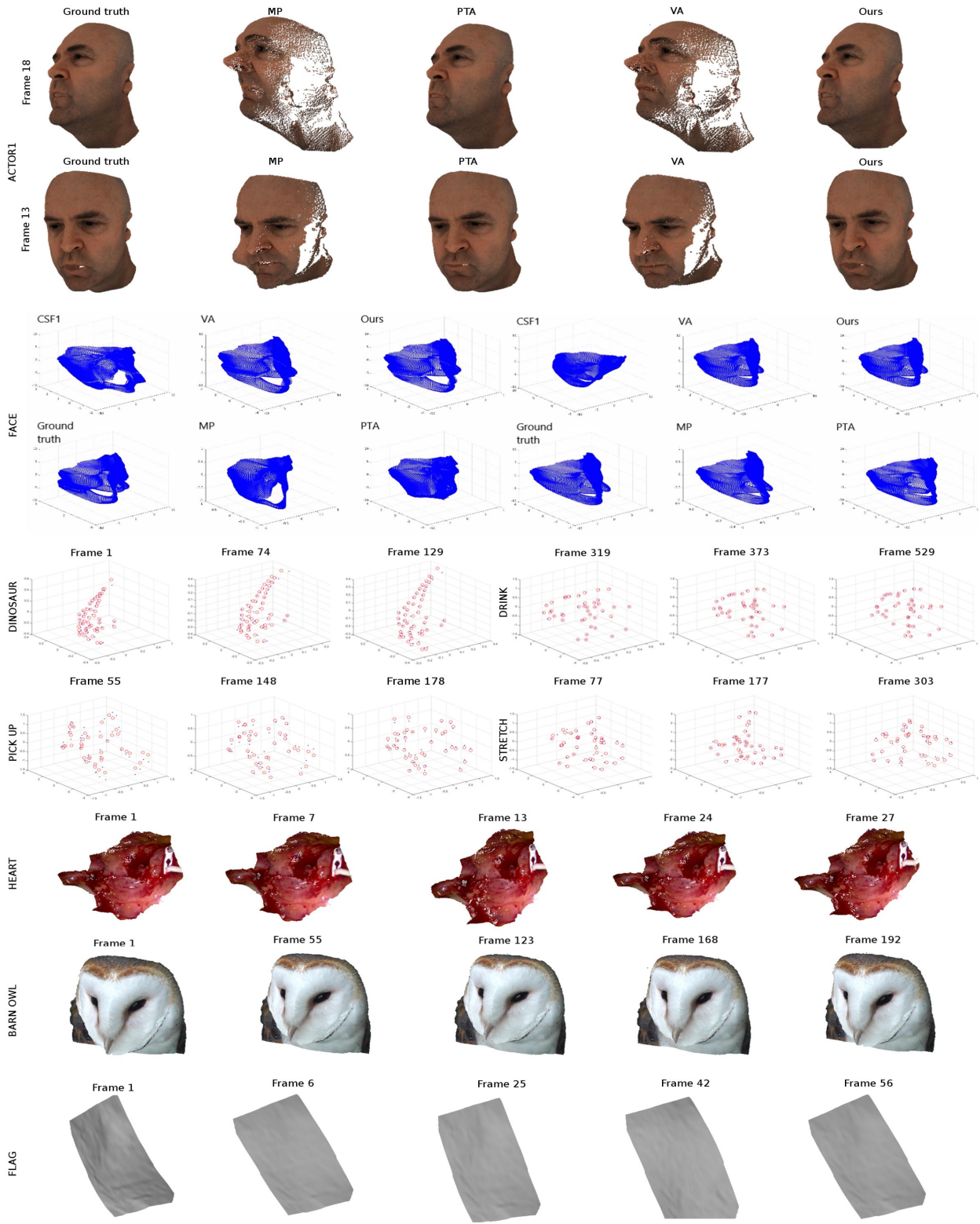

**Figure 4:** (Top 2 rows): qualitative evaluation on dense benchmark data set *Actor1* [6]; (rows 3–4): qualitative evaluation on dense benchmark data sets *Face1* (left) and *Face2* (right) [12]; (rows 5–6): qualitative evaluation on sparse benchmark data sets *Dinosaur, Pickup, Drink, Stretch* [33, 3]; (last three rows): qualitative evaluation of our approach on dense benchmark data sets *Heart* [30], *Barn Owl* [11] and *Flag* [34].